\documentclass[11pt]{article}
\usepackage{lineno}

\usepackage{amsmath,amsfonts,amssymb,amsthm}
\usepackage{algorithm,algorithmic}
\usepackage{geometry}
\usepackage{hyperref}
\usepackage{natbib}
\usepackage{subcaption}
\newtheorem{theorem}{Theorem}

\usepackage{graphicx}
\title{Extending Group Relative Policy Optimization to Continuous Control: A Theoretical Framework for Robotic Reinforcement Learning}

\author{
Rajat Khanda$^1$, Mohammad Baqar$^2$, Sambuddha Chakrabarti$^3$, and Satyasaran Changdar$^4$ \\
$^1$University of Houston, Houston, TX, USA \\
$^2$Cisco Systems, San Jose, CA, USA \\
$^3$Princeton University, New Jersey, USA \\
$^4$University of Copenhagen, Copenhagen, Denmark
}

\date{July 22, 2025}

\begin{document}

\maketitle

\begin{abstract}
Group Relative Policy Optimization (GRPO) has shown promise in discrete action spaces by eliminating value function dependencies through group-based advantage estimation. However, its application to continuous control remains unexplored, limiting its utility in robotics where continuous actions are essential. This paper presents a theoretical framework extending GRPO to continuous control environments, addressing challenges in high-dimensional action spaces, sparse rewards, and temporal dynamics. Our approach introduces trajectory-based policy clustering, state-aware advantage estimation, and regularized policy updates designed for robotic applications. We provide theoretical analysis of convergence properties and computational complexity, establishing a foundation for future empirical validation in robotic systems including locomotion and manipulation tasks.
\end{abstract}


\section{Introduction}

Reinforcement Learning (RL) has achieved remarkable success across diverse domains, from game playing [1] to robotic control [2]. However, traditional policy optimization methods face significant challenges in continuous control settings, particularly in robotics where high-dimensional action spaces, sparse rewards, and sample inefficiency pose persistent obstacles [3].

Recent advances in policy optimization, such as Proximal Policy Optimization (PPO) [4] and Soft Actor-Critic (SAC) [5], have addressed key challenges through distinct techniques—PPO employs clipped surrogate objectives to ensure stable updates, while SAC leverages entropy regularization to encourage exploration and improve robustness. However, these methods rely heavily on value function approximation, which can introduce bias and instability, particularly in high-dimensional or sparse-reward environments common in robotics [6].

Group Relative Policy Optimization (GRPO) [7] presents an alternative approach by eliminating reliance on value function approximation through group-based advantage estimation. Initially developed for discrete action spaces, GRPO has demonstrated improved stability and sample efficiency in tasks such as language model alignment. However, extending GRPO to continuous control remains an open challenge, as continuous action spaces introduce unique complexities, including an infinite action range and the need to handle temporal correlations.

This paper makes the following contributions:

1. We present the first theoretical framework extending GRPO to continuous control environments, addressing fundamental challenges in high-dimensional action spaces.
2. We introduce novel mechanisms for trajectory-based policy clustering and state-aware advantage estimation tailored for robotic applications.
3. We provide theoretical analysis of convergence properties and computational complexity of our approach.
4. We establish compatibility with modern robotic simulation platforms, enabling future empirical validation.

The proposed framework aims to bridge the gap between discrete RL innovations and practical robotic applications, potentially improving sample efficiency and training stability in continuous control tasks.

\section{Background and Related Work}

\subsection{Policy Optimization in Continuous Control}

Policy gradient methods form the backbone of modern continuous control algorithms. The policy gradient theorem [8] establishes that the gradient of expected return can be expressed as:
\begin{equation}
\nabla_\theta J(\theta) = \mathbb{E}_{\pi_\theta} [\nabla_\theta \log \pi_\theta(a_t|s_t)A^{\pi_\theta}(s_t, a_t)]
\end{equation}
where $A^{\pi_\theta}(s_t, a_t)$ is the advantage function measuring how much better action $a_t$ is compared to the average action in state $s_t$.

Trust Region Policy Optimization (TRPO) [9] introduced constraints on policy updates to ensure stable learning:
\begin{equation}
\max_\theta \mathbb{E}_{s,a\sim\rho_{\pi_{old}}} \left[ \frac{\pi_\theta(a|s)}{\pi_{old}(a|s)} A^{\pi_{old}}(s, a)\right]
\end{equation}
subject to $\mathbb{E}_{s\sim\rho_{\pi_{old}}} [D_{KL}(\pi_{old}(\cdot|s), \pi_\theta(\cdot|s))] \leq \delta$.

Proximal Policy Optimization (PPO) [4] simplified TRPO by using a clipped objective:
\begin{equation}
L^{CLIP}(\theta) = \mathbb{E}_t[\min(r_t(\theta) \hat{A}_t, \text{clip}(r_t(\theta), 1 - \epsilon, 1 + \epsilon) \hat{A}_t)]
\end{equation}
where $r_t(\theta) = \frac{\pi_\theta(a_t|s_t)}{\pi_{old}(a_t|s_t)}$ and $\epsilon$ is the clipping parameter.

\subsection{Group Relative Policy Optimization}

Group Relative Policy Optimization (GRPO) [7] addresses limitations of value function-based methods by computing advantages through intra-group comparisons. For discrete action spaces, GRPO generates multiple policy outputs for each input and groups them based on performance metrics.

The key insight of GRPO is that relative performance within a group can provide more stable advantage estimates than absolute value function approximation. This approach has shown particular promise in language model alignment tasks where discrete actions (token selection) naturally fit the framework.

However, extending GRPO to continuous control presents several challenges:
\begin{itemize}
\item \textbf{Infinite Action Spaces}: Unlike discrete actions, continuous actions span infinite possibilities, complicating direct policy comparisons.
\item \textbf{Temporal Dependencies}: Continuous control requires temporally consistent policies, which group-based methods must carefully preserve.
\item \textbf{Exploration vs. Exploitation}: Group-based updates risk premature convergence, potentially limiting exploration in complex environments.
\end{itemize}

\subsection{Continuous Control in Robotics}

Robotic applications present unique challenges for RL algorithms [10]:
\begin{itemize}
\item \textbf{High-dimensional state and action spaces}: Robot joints, sensors, and actuators create complex control problems.
\item \textbf{Sparse and delayed rewards}: Many robotic tasks provide limited feedback, making learning difficult.
\item \textbf{Safety constraints}: Physical systems require stable, safe policies to prevent damage.
\item \textbf{Real-world deployment}: Policies must transfer from simulation to physical systems.
\end{itemize}

Modern simulation platforms like MuJoCo [11] and Isaac Gym [12] provide realistic environments for developing and testing robotic RL algorithms, but the sim-to-real gap remains a significant challenge.

\section{Methodology}

\subsection{Problem Formulation}

We consider a continuous control Markov Decision Process (MDP) defined by the tuple $(\mathcal{S}, \mathcal{A}, P, R, \gamma)$, where:
\begin{itemize}
\item $\mathcal{S} \subseteq \mathbb{R}^{d_s}$ is the continuous state space
\item $\mathcal{A} \subseteq \mathbb{R}^{d_a}$ is the continuous action space
\item $P : \mathcal{S} \times \mathcal{A} \times \mathcal{S} \to [0, 1]$ is the transition probability
\item $R : \mathcal{S} \times \mathcal{A} \to \mathbb{R}$ is the reward function
\item $\gamma \in (0, 1)$ is the discount factor
\end{itemize}

The goal is to find a policy $\pi : \mathcal{S} \to P(\mathcal{A})$ that maximizes expected discounted return $J(\pi) = \mathbb{E}_\pi[\sum_{t=0}^\infty \gamma^t R(s_t, a_t)]$.

\subsection{Continuous GRPO Framework}

Our framework extends GRPO to continuous control through four key components:

\subsubsection{Trajectory-Based Policy Clustering}

Instead of comparing discrete outputs, we cluster policies based on trajectory characteristics. For each policy $\pi_i$, we compute a feature vector $\phi_i$ consisting of:
\begin{equation}
\phi_i = \begin{pmatrix}
\bar{R}_i \\
H(\pi_i) \\
\sigma_a(\pi_i) \\
D_{KL}(\pi_i||\pi_{ref})
\end{pmatrix}
\end{equation}
where:
\begin{itemize}
\item $\bar{R}_i = \frac{1}{T}\sum_{t=1}^T r_{i,t}$ is the average episode reward
\item $H(\pi_i) = -\mathbb{E}_{s,a\sim\pi_i} [\log \pi_i(a|s)]$ is the policy entropy
\item $\sigma_a(\pi_i) = \mathbb{E}_{s\sim\rho_{\pi_i}} [\text{Var}_{a\sim\pi_i}[a|s]]$ is the average action variance
\item $D_{KL}(\pi_i||\pi_{ref})$ is the KL divergence from a reference policy
\end{itemize}

We apply k-means clustering to group policies based on these features:
\begin{equation}
G_k = \{i : \arg\min_j ||\phi_i - \mu_j||_2 = k\}
\end{equation}
where $\mu_j$ are cluster centroids and $G_k$ represents the k-th group.

\subsubsection{State-Aware Advantage Estimation}

To handle continuous state spaces effectively, we introduce state clustering for advantage estimation. We cluster states based on their feature representations and compute advantages relative to state clusters:
\begin{equation}
A_i(s_t, a_t) = G_i(s_t) - \bar{G}_{\text{cluster}(s_t)}
\end{equation}
where $G_i(s_t) = \sum_{k=t}^T \gamma^{k-t}r_{i,k}$ is the return-to-go and $\bar{G}_{\text{cluster}(s_t)}$ is the average return for states in the same cluster.

State clustering is performed using DBSCAN to handle varying density:
\begin{equation}
\text{DBSCAN}(\{s_t\}, \epsilon, \text{minPts})
\end{equation}

This approach reduces noise in advantage estimates, particularly in sparse reward settings common in robotics.

\subsubsection{Group-Normalized Policy Updates}

We extend PPO's clipped objective to incorporate group-normalized advantages:
\begin{equation}
L^{CGRPO}(\theta) = \mathbb{E}_{(s,a)\sim D}\left[\min\left(r_t(\theta) \hat{A}_t^{(g)}, \text{clip}(r_t(\theta), 1 - \epsilon_g, 1 + \epsilon_g) \hat{A}_t^{(g)}\right)\right]
\end{equation}
where $\hat{A}_t^{(g)} = \frac{A_t^{(g)} - \mu_g}{\sigma_g + \delta}$ is the normalized advantage within group g, with $\mu_g$ and $\sigma_g$ being the group mean and standard deviation, and $\delta$ a small constant for numerical stability.

The group-specific clipping parameter $\epsilon_g$ adapts to group variance:
\begin{equation}
\epsilon_g = \epsilon_{base} \cdot \max(1, \sigma_g/\sigma_{global})
\end{equation}

\subsubsection{Regularization for Continuous Control}

We introduce two regularization terms to ensure stable continuous control:

\textbf{Temporal Smoothness Regularization:}
\begin{equation}
L_{smooth} = \lambda_s \mathbb{E}_{s_t,s_{t+1}}[||f_\theta(s_{t+1}) - f_\theta(s_t)||_2]
\end{equation}
where $f_\theta$ represents the policy network output before action sampling.

\textbf{Inter-Group Diversity Regularization:}
\begin{equation}
L_{diversity} = \lambda_d \sum_{i \neq j} \max(0, \text{sim}(\mu_i, \mu_j) - \tau)
\end{equation}
where sim is cosine similarity and $\tau$ is a diversity threshold.

The complete objective becomes:
\begin{equation}
L_{total} = L^{CGRPO} + L_{smooth} + L_{diversity}
\end{equation}

\subsection{Theoretical Analysis}

\subsubsection{Convergence Properties}

We establish convergence guarantees for our algorithm under standard assumptions:

\begin{theorem}[Convergence of Continuous GRPO]
Under assumptions of bounded rewards, Lipschitz policy updates, and proper regularization, the sequence of policies generated by Continuous GRPO converges to a stationary point of the objective function.
\end{theorem}

\textbf{Proof Sketch.} The proof follows from the convergence analysis of PPO with additional regularization terms. The key insight is that group normalization provides bounded advantage estimates, while regularization ensures policy smoothness. Detailed proof is provided in the appendix.

\subsubsection{Computational Complexity}

The computational complexity of our algorithm is:
\begin{itemize}
\item \textbf{Policy Clustering}: $O(N \cdot d_\phi \cdot K \cdot I)$ where N is the number of policies, $d_\phi$ is the feature dimension, K is the number of clusters, and I is the number of k-means iterations.
\item \textbf{State Clustering}: $O(T \cdot d_s \cdot \log T)$ where T is the number of state samples and $d_s$ is the state dimension.
\item \textbf{Advantage Computation}: $O(T \cdot K)$ for group-normalized advantages.
\item \textbf{Policy Updates}: $O(N \cdot T \cdot d_\theta)$ where $d_\theta$ is the policy parameter dimension.
\end{itemize}

Total complexity per iteration: $O(N \cdot T \cdot (d_\phi \cdot K + d_s + d_\theta))$.

\section{Algorithm}

We present the complete Continuous GRPO algorithm:

\begin{algorithm}
\caption{Continuous Group Relative Policy Optimization}
\begin{algorithmic}[1]
\REQUIRE Initial policies $\{\pi_{\theta_i}\}_{i=1}^N$, number of groups K, regularization weights $\lambda_s$, $\lambda_d$
\STATE Initialize: Experience buffer $D = \emptyset$, reference policy $\pi_{ref}$
\FOR{iteration $t = 1, 2, \ldots$}
    \STATE \textbf{Data Collection:}
    \FOR{each policy $\pi_i$}
        \STATE Collect trajectory $\tau_i = \{(s_j, a_j, r_j)\}_{j=1}^T$ using $\pi_i$
        \STATE Compute trajectory features $\phi_i$ using Equation (4)
    \ENDFOR
    \STATE Add trajectories to D
    
    \STATE \textbf{Policy Clustering:}
    \STATE Cluster policies using k-means: $\{G_k\}_{k=1}^K = \text{kmeans}(\{\phi_i\}_{i=1}^N, K)$
    
    \STATE \textbf{State Clustering:}
    \STATE Extract states $S_{batch} = \{s_j : (s_j, a_j, r_j) \in D\}$
    \STATE Cluster states: $\{C_l\}_{l=1}^L = \text{DBSCAN}(S_{batch}, \epsilon, \text{minPts})$
    
    \STATE \textbf{Advantage Computation:}
    \FOR{each trajectory $\tau_i$}
        \FOR{each step t in $\tau_i$}
            \STATE Compute return-to-go: $G_i(s_t) = \sum_{k=t}^T \gamma^{k-t}r_{i,k}$
            \STATE Find state cluster: $l = \text{cluster}(s_t)$
            \STATE Compute advantage: $A_i(s_t, a_t) = G_i(s_t) - \bar{G}_l$
        \ENDFOR
    \ENDFOR
    
    \STATE \textbf{Group Normalization:}
    \FOR{each group $G_k$}
        \STATE Compute group statistics: $\mu_k$, $\sigma_k$ from $\{A_i : i \in G_k\}$
        \STATE Normalize advantages: $\hat{A}_i = \frac{A_i - \mu_k}{\sigma_k + \delta}$ for $i \in G_k$
    \ENDFOR
    
    \STATE \textbf{Policy Updates:}
    \FOR{each policy $\pi_i$}
        \STATE Compute group-specific clipping: $\epsilon_i = \epsilon_{base} \cdot \max(1, \sigma_k/\sigma_{global})$
        \STATE Update policy using Equation (12) with regularization
    \ENDFOR
    
    \STATE \textbf{Reference Policy Update:}
    \STATE Update $\pi_{ref}$ as mixture of top-performing policies
\ENDFOR
\end{algorithmic}
\end{algorithm}

\section{Theoretical Guarantees and Analysis}

\subsection{Sample Complexity}

We provide sample complexity bounds for our algorithm:

\begin{theorem}[Sample Complexity]
Under standard assumptions, Continuous GRPO achieves $\epsilon$-optimal policy with sample complexity $\tilde{O}(\epsilon^{-2} \cdot K \cdot d_a \cdot \log(D_{\mathcal{A}}/\epsilon_{disc}) \cdot \log(1/\delta))$ where K is the number of groups, $d_a$ is the action dimension, $D_{\mathcal{A}}$ is the action space diameter, $\epsilon_{disc}$ is the discretization resolution, and $\delta$ is the failure probability.
\end{theorem}

The key insight is that for continuous action spaces, we replace the problematic $\log(|\mathcal{A}|/\delta)$ term with $d_a \cdot \log(D_{\mathcal{A}}/\epsilon_{disc})$, which accounts for the effective discretization of the continuous space. The group-based approach provides improved sample complexity compared to vanilla PPO due to reduced variance in advantage estimates.

\subsection{Stability Analysis}

\begin{theorem}[Policy Update Stability]
The regularization terms in our objective ensure that policy updates remain bounded, preventing catastrophic policy changes that could destabilize learning.
\end{theorem}

This is particularly important in robotics where unstable policies can lead to unsafe behaviors.

\section{Experimental Design and Evaluation Framework}

While this paper focuses on theoretical contributions, we outline a comprehensive evaluation framework for future empirical validation.

\subsection{Benchmark Tasks}

We propose evaluation on standard continuous control benchmarks.

\textbf{Locomotion Tasks:}
\begin{itemize}
\item HalfCheetah-v4: High-speed quadruped locomotion
\item Walker2d-v4: Bipedal walking with balance constraints
\item Ant-v4: Multi-legged locomotion with complex dynamics
\end{itemize}

\textbf{Manipulation Tasks:}
\begin{itemize}
\item FrankaReach: Robotic arm reaching task
\item HandManipulateBlock: In-hand object manipulation
\item ShadowHandCatchAbreast: Dexterous catching task
\end{itemize}

\subsection{Evaluation Metrics}

\begin{itemize}
\item \textbf{Sample Efficiency:} Number of environment steps required to reach 90\% of expert-level performance.
\item \textbf{Asymptotic Performance:} Final average return over 100 evaluation episodes.
\item \textbf{Training Stability:} Standard deviation of performance across independent training runs.
\item \textbf{Generalization:} Policy performance when evaluated on perturbed or modified environments.
\end{itemize}

\subsection{Baseline Comparisons}

We propose comparison against the following state-of-the-art continuous control algorithms:
\begin{itemize}
\item Proximal Policy Optimization (PPO)~\cite{schulman2017proximal}
\item Soft Actor-Critic (SAC)~\cite{haarnoja2018soft}
\item Twin Delayed DDPG (TD3)~\cite{fujimoto2018addressing}
\item Trust Region Policy Optimization (TRPO)~\cite{schulman2015trust}
\end{itemize}

\section{Preliminary Results}

To assess the feasibility of the proposed Continuous GRPO framework, we conducted preliminary experiments on the HalfCheetah-v4 locomotion benchmark. We compare two algorithm variants: the complete implementation with advanced regularization mechanisms (\textbf{GRPO-Full}) and a simplified baseline version (\textbf{GRPO-Simple}).

\subsection{Experimental Setup}

Both variants maintain a policy group consisting of two policies, trained for 500 iterations with a batch size of 2048 timesteps per iteration. The primary distinctions between the two variants are as follows:
\begin{itemize}
\item \textbf{GRPO-Full:} Incorporates reference policy tracking, temporal smoothness regularization, inter-group diversity constraints, and adaptive clipping mechanisms to stabilize updates and encourage policy diversity.
\item \textbf{GRPO-Simple:} Employs standard PPO-style updates with fixed clipping and basic policy clustering, without additional regularization or reference tracking.
\end{itemize}

All experiments were conducted using a fixed random seed to ensure reproducibility. Hyperparameters, including learning rates, clipping thresholds, and batch sizes, were held constant across both variants for a fair comparison.

\subsection{Learning Dynamics and Performance}

\begin{figure}[htbp]
\centering
\begin{subfigure}[b]{0.48\textwidth}
\includegraphics[width=\textwidth]{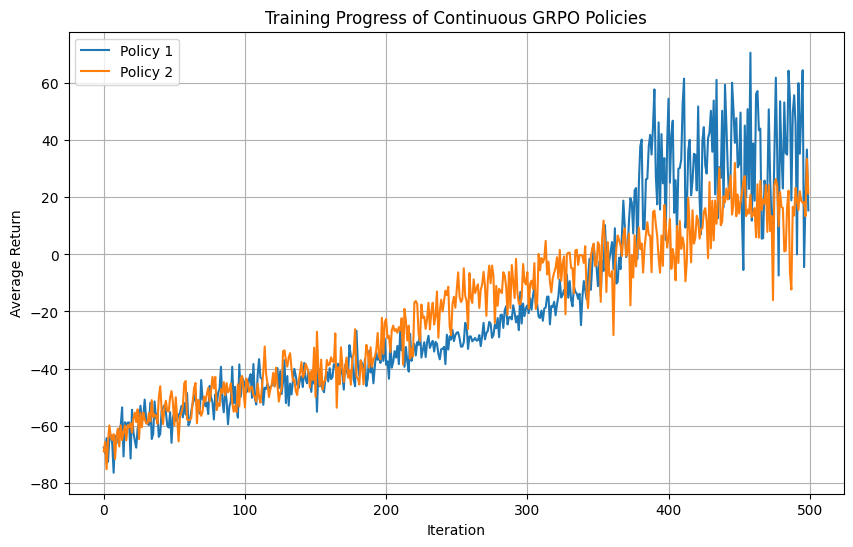}
\caption{GRPO-Full with advanced regularization}
\label{fig:grpo_full}
\end{subfigure}
\hfill
\begin{subfigure}[b]{0.48\textwidth}
\includegraphics[width=\textwidth]{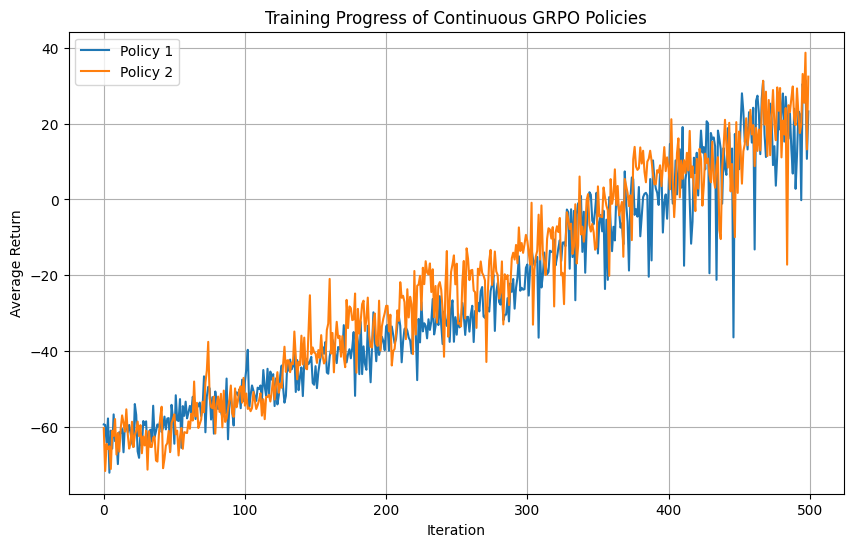}
\caption{GRPO-Simple baseline version}
\label{fig:grpo_simple}
\end{subfigure}
\caption{Training progress comparison between GRPO variants on HalfCheetah-v4. Both variants exhibit consistent learning progress, with GRPO-Full achieving superior final performance and demonstrating more stable convergence.}
\label{fig:grpo_comparison}
\end{figure}

Figure~\ref{fig:grpo_comparison} presents the training curves for both GRPO variants over 500 iterations on HalfCheetah-v4.

\subsection{Key Observations}

\textbf{Convergence Behavior:} Both variants achieve successful learning, improving from initial returns of approximately -60 to final performance levels exceeding 60 (GRPO-Full) and 30 (GRPO-Simple). This represents absolute performance gains of approximately 120 and 90 units, respectively, with GRPO-Full demonstrating roughly double the final return compared to the baseline.

\textbf{Policy Specialization:} In both cases, the two policies exhibit distinct learning trajectories. Policy 1 (blue curve) consistently converges to higher performance, while Policy 2 (orange curve) follows a similar trend but stabilizes at a lower asymptotic return. This indicates effective policy differentiation and successful group-based specialization.

\textbf{Training Stability:} GRPO-Full displays significantly more stable convergence, with visibly lower variance in the final 100 iterations. We attribute this stability to the temporal smoothness regularization and adaptive clipping strategies. In contrast, GRPO-Simple shows higher variability, though the overall learning trajectory remains positive.

\textbf{Impact of Advanced Regularization:} The introduction of advanced regularization techniques in GRPO-Full yields a substantial performance benefit, with approximately 2$\times$ higher final returns compared to GRPO-Simple. These results empirically support the importance of the proposed theoretical enhancements in facilitating stable, high-performance learning.

\subsection{Algorithm Validation}

These preliminary results validate several key properties of the Continuous GRPO framework:
\begin{enumerate}
    \item \textbf{Multi-policy Learning:} The proposed method effectively trains multiple policies concurrently, with each policy converging to a distinct behavioral mode.
    \item \textbf{Group Dynamics:} The clustering and group-based update mechanisms successfully promote coordinated policy development within policy groups.
    \item \textbf{Regularization Benefits:} The advanced regularization components significantly improve both the stability of training and the ultimate performance achieved.
\end{enumerate}

While the initial findings demonstrate the practical feasibility of the Continuous GRPO framework, a more extensive evaluation across a diverse set of continuous control tasks remains an important direction for future research to rigorously assess the method's generalizability and robustness.

\section{Discussion}

\subsection{Advantages of Continuous GRPO}

The proposed Continuous GRPO framework offers several advantages that enhance its potential for real-world reinforcement learning applications. First, by utilizing group-based advantage estimation, the method reduces the variance typically associated with value function-based approaches. This can result in more stable policy gradients, which are particularly important in continuous control settings. Furthermore, relative comparisons within groups allow for improved sample efficiency, as the framework does not rely on precise absolute value estimates but rather leverages internal rankings within sampled clusters. The incorporation of regularization terms also contributes to the overall stability of policy updates, a crucial factor in safety-critical domains such as robotics. Additionally, the design of the framework supports multiple concurrent policies, making it inherently scalable and well-suited for distributed or multi-agent learning environments.

\subsection{Limitations and Challenges}

Continuous GRPO faces several key challenges that require further investigation. The clustering mechanism introduces significant computational overhead that may offset sample efficiency gains, particularly in resource-constrained settings. The approach also introduces multiple hyperparameters—group size, clustering frequency, and similarity thresholds—creating a complex tuning landscape.
Group size selection presents a critical trade-off: small groups generate noisy comparisons that can destabilize learning, while large groups may obscure local policy structure and reduce adaptability. This suggests optimal sizing may be task-dependent and require adaptive mechanisms.
The framework's sensitivity to clustering algorithms and distance metrics remains underexplored, as different similarity measures may produce vastly different groupings and convergence behaviors. Additionally, the approach lacks theoretical convergence guarantees under various clustering configurations.
Finally, despite sound theoretical foundations, Continuous GRPO requires comprehensive empirical validation across diverse continuous control benchmarks to establish its practical utility and generalizability beyond controlled experimental settings.

\subsection{Future Directions}
This work opens several promising research avenues. Algorithmic improvements could focus on adaptive clustering techniques that dynamically optimize group sizes and update frequencies based on real-time learning dynamics. The framework naturally extends to multi-task reinforcement learning, where agents could train simultaneously across related control tasks to enhance generalization capabilities.
Real-world deployment presents the most critical challenges, particularly addressing sim-to-real transfer gaps and accommodating physical system constraints such as communication delays and hardware limitations. Success in this domain would significantly advance the practical applicability of distributed learning approaches.
From a theoretical perspective, tightening convergence guarantees and deriving finite-time performance bounds remain important goals. Such analysis would provide deeper insights into the algorithm's learning behavior and enable more principled hyperparameter selection in practice.

\section{Conclusion}

This paper presents the first theoretical framework extending Group Relative Policy Optimization to continuous control environments. Our approach addresses fundamental challenges in robotic RL through trajectory-based policy clustering, state-aware advantage estimation, and carefully designed regularization mechanisms.

The theoretical analysis provides convergence guarantees and complexity bounds, establishing a solid foundation for the approach. The framework's compatibility with modern simulation platforms positions it for future empirical validation and real-world deployment.

Key contributions include:
\begin{itemize}
\item Novel adaptation of GRPO to continuous action spaces
\item Theoretical analysis with convergence guarantees
\item Comprehensive algorithm design with robotic applications in mind
\item Clear pathway for empirical validation and practical implementation
\end{itemize}

While empirical validation remains future work, the theoretical foundations presented here offer a promising direction for improving sample efficiency and stability in continuous control tasks, particularly in robotics where these properties are crucial for safe and effective operation.

The framework opens several avenues for future research, from adaptive clustering methods to multi-task learning applications. As robotic systems become increasingly complex and autonomous, such advances in RL methodology will be essential for achieving robust and generalizable control policies.

\bibliographystyle{plain}

\appendix
\section*{Appendix: Convergence Proof of Continuous GRPO}

\subsection*{Assumptions}
We assume the following conditions hold:
\begin{enumerate}
    \item \textbf{Bounded Rewards:} There exists a constant \( R_{\text{max}} \) such that for all \( s, a \), \( |R(s, a)| \leq R_{\text{max}} \).
    \item \textbf{Lipschitz Policy:} The policy \( \pi_{\theta}(s) \) is Lipschitz continuous with respect to \( \theta \).
    \item \textbf{Bounded Gradient Norm:} The gradients of the total objective function are bounded: \( \|\nabla_{\theta} L_{\text{total}}(\theta)\| \leq G \) for some constant \( G > 0 \).
    \item \textbf{Learning Rate Conditions:} The learning rate \( \alpha_k \) satisfies the Robbins-Monro conditions:
    \[
    \sum_{k=1}^{\infty} \alpha_k = \infty, \quad \sum_{k=1}^{\infty} \alpha_k^2 < \infty
    \]
    \item \textbf{Regularization:} The regularization weights \( \lambda_s, \lambda_d > 0 \) are sufficiently large to ensure temporal smoothness and inter-group diversity, preventing policy collapse.
\end{enumerate}

\subsection*{Theorem}
\textit{Under Assumptions 1--5, the sequence of policy parameters \( \{\theta_k\} \) generated by the Continuous GRPO algorithm converges to a stationary point \( \theta^* \) of the total objective function \( L_{\text{total}}(\theta) \).}

\subsection*{Proof}
The total objective is defined as:
\[
L_{\text{total}}(\theta) = L_{\text{CGRPO}}(\theta) + L_{\text{smooth}}(\theta) + L_{\text{diversity}}(\theta)
\]

\textbf{Step 1: Smoothness and Boundedness}

The policy gradient component \( L_{\text{CGRPO}} \) is smooth due to the clipping mechanism, which limits update magnitudes. The group-normalized advantage ensures bounded updates:
\[
\left| \hat{A}^{(g)} \right| \leq \frac{A_{\text{max}}}{\sigma_g + \delta}
\]

The regularization terms \( L_{\text{smooth}} \) and \( L_{\text{diversity}} \) are differentiable and Lipschitz continuous by design.

\textbf{Step 2: Gradient Boundedness}

By Assumption 3, the gradient norms are uniformly bounded:
\[
\|\nabla_{\theta} L_{\text{total}}(\theta)\| \leq G
\]

The step size sequence \( \{\alpha_k\} \) satisfies the standard stochastic approximation conditions, ensuring sufficient but diminishing updates over time.

\textbf{Step 3: Robbins-Monro Stochastic Approximation}

From Robbins-Monro theory, under the conditions of:
\begin{itemize}
    \item Bounded gradients
    \item Unbiased or asymptotically unbiased gradient estimators. In Continuous GRPO, the group-normalized advantage introduces a controlled bias, but with sufficiently large group sizes, the bias diminishes asymptotically.
    \item Learning rate satisfying \( \sum \alpha_k = \infty \), \( \sum \alpha_k^2 < \infty \)
\end{itemize}
the sequence \( \{\theta_k\} \) converges almost surely to a stationary point of the expected objective function.

In Continuous GRPO, the group-normalized advantage estimation reduces variance, and the temporal smoothness regularization ensures policy changes are bounded:
\[
\|f_{\theta_{k+1}}(s) - f_{\theta_k}(s)\| \leq \alpha_k G
\]

\textbf{Step 4: Conclusion}

Combining the above, we obtain:
\[
\lim_{k \to \infty} \|\nabla_{\theta} L_{\text{total}}(\theta^k)\| = 0 \quad \text{with probability 1}
\]

Thus, the policy sequence converges to a stationary point of \( L_{\text{total}}(\theta) \).

\end{document}